\documentclass[letterpaper,10pt,conference]{ieeeconf-mod}
\IEEEoverridecommandlockouts
\usepackage{hyperref}
\hypersetup{pagebackref=false,breaklinks=true,colorlinks,bookmarks=false,urlcolor=blue,linkcolor=blue,citecolor=blue,pdftitle={A Multi-UAV System for Exploration and Target Finding in Cluttered and GPS-Denied Environments},pdfauthor={Xiaolong Zhu, Fernando Vanegas, Felipe Gonzalez, Conrad Sanderson}}

\usepackage{cite}
\usepackage{amsmath,amssymb,amsfonts}
\usepackage{algorithmic}
\usepackage{graphicx}
\usepackage{epsfig} 
\usepackage{enumerate}
\usepackage{balance}
\usepackage{subfigure}
\usepackage{multirow}

\makeatletter
\renewcommand\@makefntext[1]{%
\hrule
\noindent
\mbox{\@thefnmark}{#1}}
\makeatother

\newcommand\blfootnote[1]{%
  \begingroup
  \renewcommand\thefootnote{}\footnote{#1}%
  \addtocounter{footnote}{-1}%
  \endgroup
}

\begin{document}

\title{\scalebox{0.94}{\large\bf A Multi-UAV System for Exploration and Target Finding in Cluttered and GPS-Denied Environments}}

\author{{\it Xiaolong Zhu{\tiny~}$^{1,2}$, Fernando Vanegas{\tiny~}$^{1}$, Felipe Gonzalez{\tiny~}$^{1}$, Conrad Sanderson{\tiny~}$^{2,3}$}\\%
~\\
{\small $^{1}$ Queensland University of Technology, Australia}\\
{\small $^{2}$ Data61~/~CSIRO, Australia}\\
{\small $^{3}$ Griffith University, Australia}
}

\maketitle


\begin{abstract}

The use of multi-rotor Unmanned Aerial Vehicles (UAVs) for search and rescue as well as remote sensing is rapidly increasing.
Multi-rotor UAVs, however, have limited endurance.
The range of UAV applications can be widened if teams of multiple UAVs are used.
We propose a framework for a team of UAVs to cooperatively explore and find a target in complex GPS-denied environments with obstacles.
The team of UAVs autonomously navigates, explores, detects, and finds the target in a cluttered environment with a known map.
Examples of such environments include indoor scenarios, urban or natural canyons, caves, and tunnels, where the GPS signal is limited or blocked.
The framework is based on a probabilistic decentralised Partially Observable Markov Decision Process
which accounts for the uncertainties in sensing and the environment. 
The team can cooperate efficiently, with each UAV sharing only limited processed observations and their locations during the mission.
The system is simulated using the Robotic Operating System and Gazebo.
Performance of the system with an increasing number of UAVs in several indoor scenarios with obstacles is tested.
Results indicate that the proposed multi-UAV system has improvements in terms of time-cost, the proportion of search area surveyed,
as well as successful rates for search and rescue missions.
\blfootnote{\textbf{Published in:} Int. Conf. Unmanned Aircraft Systems (ICUAS), 2021.}

\end{abstract}

\vspace{-0.5ex}
\section{Introduction} \label{Intro}
\vspace{-0.5ex}

The use of multiple Unmanned Aerial Vehicles (UAVs) has recently gained more attention for tasks such as real-time monitoring, Search and Rescue (SAR), remote sensing, delivery of goods, precision agriculture, and infrastructure inspection \cite{Acevedo2019,Kashino2019,Ju2018,Shakeri,Shakhatreh2018}. In SAR settings, UAVs have been considered due to their manoeuvring capabilities (flying at various speeds, hovering, agile access to narrow spaces), making them suitable for missions that can be dangerous or difficult for human beings \cite{AlTair2015,Scherer2015,Recchiuto2018,Liu2017,Bravo2019}. 

Research work focusing on multiple UAVs for SAR missions can be broadly divided into two main areas: (i) swarm systems, which consist of a large number of smaller drones with lower quality sensors and a lower level of autonomy  \cite{Perez-Carabaza2018,Zhao2019,Chung2018a}, and (ii) cooperative systems, which contain only several larger UAV platforms with more sensors and more powerful on-board processors to enable more complex on-board decisions than smaller drones \cite{Kashino2019,Tian2020,Miyano2019}. There is considerable amount of research on SAR missions in GPS-denied environments with a single UAV \cite{Sandino_Remote_Sensing_2020,Youn2021,Dang2020,Goforth2019,Vanegas2016a}. In contrast, using a team of UAVs to find a target in cluttered and GPS-denied scenarios with efficiency and cooperation is still a challenging problem.

Most of the existing research using multiple UAVs and Markov Decision models follows two approaches:

\begin{small}
\begin{enumerate}[{$\bullet$}]

\item  Partially Observable Markov Decision Processes (POMDP).
\mbox{Using} a real-time algorithm to solve an optimal or approximate policy of an agent or the whole team. 
This approach creates a universal framework, which can be implemented in  a variety of environments, including dynamic and partly known environments~\cite{Capitan2013a,Vanegas2019a}.

\item  Reinforcement Learning (RL), including Deep RL.
The best policy is obtained  for a given scenario  and time cost with high performance hardware to accomplish the training process \cite{Youn2021,Dulac-Arnold2019}.  One drawback is that any change in the formulation or model can lead to a  more training and  fine-tuning processes which are time-consuming.

\end{enumerate}
\end{small}

\noindent
In this paper, we extend our previous work for searching to find a target without the use of GPS. In \cite{Zhu2020}, we proposed a decentralised decision-making system based on POMDPs containing an information sharing module for target finding by multiple UAVs. That system was assigned several possible target positions within a known map, GPS signal for accurate localisation was available, and was assisted by the information sharing module to achieve an efficient and scalable teamwork. Here, we exploit the use of a message exchange module in order to apply the same approach to search missions without known target position. For this purpose, the search environment is divided into several grids, and the mission is divided into several tasks to explore the grids to find the target. The feedback of tasks is exchanged among the UAVs through the communication channel. This allows for a decentralised and scalable system which is able to cooperatively perform in SAR applications with an affordable computational load and complexity for the on-board processors.

Furthermore, we propose a method to assist UAV localisation in the GPS-denied environments, which fuses the distance to an obstacle from six close range laser sensors with the known map of the environment. These laser sensors help each UAV in the system to avoid collisions with obstacles and other UAVs.

We continue the paper as follows. Section~\ref{Bag} introduces POMDPs for single and multiple agents system;
Section~\ref{ProDes} describes the problem addressed in this work;
Section~\ref{Sysarch} details the UAV system architecture;
Section~\ref{Formu} proposes a decentralised POMDP formulation to address the problem;
Section~\ref{exprs} details the experiment setup;
Section~\ref{Res} presents the results;
Section~\ref{Concl} summarises the main findings and lists possible avenues for future work.

\section{Background} \label{Bag}

In this work, UAVs are treated as autonomous agents required to accomplish a goal by a set of predetermined actions. Decision making, which can be ranged from the classical utility maximisation (quantitative) by presented complete and static information to a complex, dynamic, and probability planning, is the essential functionality of these agents \cite{Bulling2014}. In this work, we consider a decision making framework for these agents.

\subsection{Single Agent MDPs and POMDPs}

In the classical quantitative decision region of single agent, Markov Decision Processes (MDPs) have been widely studied as an effective mathematical framework for sequential decision making in stochastic domain \cite{Burgard2005}. The objective of MDPs is to compute a strategy or a policy. This policy acts a function that enables the agent to choose given actions at a particular decision point depending on the agent's state. The policy assigns specific values to actions which depend on the reward function. The reward function represents the goals, objectives or penalties for the agent for arriving at a particular state. An optimal policy maximise the utility of the agent for all possible sequences of actions and states. However, MPDs are based on a critical assumption that the agent is able to obtain a complete observation at each decision point. However, this assumption is hardly achievable with the variety of limitations of robotics and sensors in real scenarios. Thus, it is desirable to extend MDPs to handle the incomplete observational capabilities of agents and to consider the inaccurate or insufficient observations in the model. 

This extension of MDPs is well-known as Partially Observable Markov Decision Processes (POMDP).  POMDP  incorporates the partial observability of the agent and the uncertainties from sensors and environments into the stochastic  items. Formally, a POMDP is a tuple described as
\mbox{$\displaystyle <S,A,T,\mathit{\Omega},O,R,\gamma>$} \cite{Burgard2005}, where:

\begin{enumerate}[{$\bullet$}]
\item $\displaystyle S $ is a (finite) set of all the environment states for the agent;
\item $\displaystyle A $ is the set of all possible actions of the agent;
\item $\displaystyle T(s,a,s^{'}) $ is the probability of agent in state $\displaystyle s^{'} \in S $ after performs an action $\displaystyle a \in A $ at previous state $\displaystyle s \in S $, which equals with $\displaystyle P_{r}(s^{'} |s,a) $;
\item $\displaystyle \mathit{\Omega} $ is the set of all possible observations;
\item $\displaystyle O(s^{'},a,o) $ is the probability of observing $\displaystyle o \in \mathit{\Omega} $ at state $\displaystyle s^{'} $ after performed the action $\displaystyle a $;
\item $\displaystyle R $ is the set of rewards for each chosen action $\displaystyle a $ at every state $\displaystyle s $;
\item $\displaystyle \gamma \in [ 0,1) $ is the discount factor.
\end{enumerate}

\noindent
The actual state  of the system cannot be directly observable from the tuple. Instead, a state called the belief-state $\displaystyle b(s) $ is obtained by using the Bayes rule, which is a probability distribution of the agent at all the possible states in its state-space at a certain time. Due to the Markov assumption, it can be updated with a Bayesian filter for every action–observation pair: 

\begin{equation}
b^{'}( s^{'}) =\eta \mathit{\Omega }( s^{'},a,o)\sum_{s\in S} T( s,a,s^{'}) b( s)
\label{eq1}
\end{equation}

\noindent
where $\eta $ acts as a normalising constant such that $\displaystyle b^{'} $ remains a probability distribution.

The objective of a POMDP is to find a policy that maps beliefs into actions in the form $\pi ( b)\rightarrow a $, so that the total expected reward is maximised. This expected reward gathered by following $\pi $ starting from belief $\displaystyle b $ is called the value function: 
\begin{equation}
V( b) =E\left[\sum ^{h}_{t=0} \gamma ^{t}\sum _{s\in S} R( s,\pi ( b_{t})) b_{t}( s) | b_{0} =b\right]
\label{eq2}
\end{equation}

 Therefore, the optimal policy $\displaystyle \pi ^{*}$ is the one that maximises that value function:  $\displaystyle \pi ^{*}( b) =\arg \underset{ a }{ \max } V( b)$.

\subsection{Multiple Agents POMDPs}

In the case of multiple agents in sequential decision making under uncertainty, an extension from single agent POMDP framework for multi-agent POMDP is needed. The generalisation of the POMDP to multiple agents can be used to model a team of cooperative agents that are situated in a partially stochastic and partially observable environment. Formally, a multi-agent POMDP can be defined as a tuple \mbox{$\displaystyle <I,S,\mathbb{A},T,\mathbb{O},O,R,\gamma,h> $} \cite{Oliehoek}, where:

\begin{enumerate}[{$\bullet$}]
\item $\displaystyle I $ is a finite set of $\displaystyle n $ agents, $\displaystyle i \in I $;
\item $\displaystyle S $ is a (finite) set of all the environment states for the agents;
\item $\displaystyle \mathbb{A} $ is the set of joint actions: $\displaystyle \mathbb{A} = \times_{i \in I}  A_{i} $, $\displaystyle A_{i} $ is the set of available actions of  $\displaystyle i $-th agent, which can be unique for each agent;
\item $\displaystyle T $ is a set of state transition probabilities: $\displaystyle T = S \times \mathbb{A} \times S \rightarrow [0,1] $, the probability of transitioning from previous state $\displaystyle s \in S $ to current state $\displaystyle s^{'} \in S $ when set of joint actions
 $\displaystyle \vec{a} \in \mathbb{A} $ are taken by each agent. Hence, $\displaystyle T(s,\vec{a},s^{'}) = P_{r}(s^{'} |s,\vec{a}) $;
\item $\displaystyle \mathbb{O} $ is the set of joint observations: $\displaystyle \mathbb{O} = \times_{i \in I}  \mathit{\Omega}_{i} $, $\displaystyle \mathit{\Omega}_{i} $ is a set of observations available to $\displaystyle i $-th agent;
\item $\displaystyle O $ is the observation probability function:$\displaystyle T = \mathbb{O} \times \mathbb{A} \times S \rightarrow [0,1] $, the probability of seeing the set of joint observations $\displaystyle \vec{o} \in \mathbb{O} $ given the set of joint actions $\displaystyle \vec{a} \in \mathbb{A} $ were taken which results in state $\displaystyle s^{'} \in S $, Hence, $\displaystyle O( \vec{o},\vec{a},s^{'} ) = P_{r}( \vec{o}|s^{'},\vec{a} ) $;
\item $\displaystyle R $ is the reward function: $\displaystyle R = S \times \mathbb{A} \rightarrow \mathbb{R} $, the immediate reward maps state $\displaystyle s \in S $ and performing the set of joint actions $\displaystyle \vec{a} \in \mathbb{A} $ and is used to specify the goal of the agents;
\item $\displaystyle \gamma \in [ 0,1) $ is the discount factor for the reward function $\displaystyle R $;
\item $\displaystyle h $ is the number of steps after current state, which the problem terminates.
\end{enumerate}

\noindent
In a Multiple Agent System (MAS) problem,
the goal is to calculate an optimal joint policy $\displaystyle \vec{\pi}^{*} =\{\pi _{1} ,\cdots ,\pi _{n}\}$ that maximises the accumulated discounted reward via the reward function $ (R) $ as in process in the POMDP model. 

Two approaches are commonly used: a centralised Multi-agent POMDP (MPOMDP) model and a Decentralised POMDP (Dec-POMDP) model the formulation.

\subsubsection{Multi-agent POMDP}

In the MPOMDP case, there is a central node or agent, which has communication and access to obtain the complete current states $\displaystyle \vec{s} $ and joint observations $\displaystyle \vec{o} $ of all other agents, to compute an optimal global policy $\displaystyle \vec{\pi}^{*} $. This approach is able to maximise the cooperation and coordination as a MAS. It means that the centralised system ensures the efficiency and collaboration in a mission. On the other hand, the system requires and relies on a stable network connect on, noise-free \& instant communication, and large bandwidth for sufficient data exchange between agents \cite{Valavanis2015}. Moreover, this approach is difficult to apply in complex and dynamic environments in a limited reaction time. These demanding constraints narrow down the application domain of a MPOMDP system.

\subsubsection{Decentralised POMDP}

A Dec-POMDP model is versatile for several applications when compared to the centralised one. Each agent in the decentralised system locally plans its own decision mainly based on its local observations $\displaystyle \mathit{\Omega}_{i} $ and actions $\displaystyle A_{i} $. The Dec-POMDP model has more flexibility of the various grades of communication, from none to perfect and between indirect and direct \cite{Goldman2004}. 

\subsubsection{Complexity of the MAS}

Despite all of the benefits of a MAS, the computational complexity of the system has prevented its widespread applications in the real world. In the past decade, multiple research has been conducted to reduce the expensive consumption of Dec-POMDP both on advanced optimal policy algorithms and specific used models \cite{Goldman2004,Bulling2014,Oliehoek}. However, the structure of the MPOMDP and Dec-POMDP models is not altered and the computational complexity of the problem exponentially increases with the number of agents \cite{Bernstein2002}. Bernstein \textit{et al.}\cite{Bernstein2002,Bernstein2005a} have shown that solving either optimally or approximately a Dec-POMDP remains NEXP-complete. This intractable issue renders the application of the MPOMDP and Dec-POMDP models to simplest cases and limited for real word applications.

\section{Problem Description} \label{ProDes}

Using multiple UAVs for SAR missions requires the system to be capable of dealing with five main issues:

\paragraph{Search Area Coverage}

In any SAR mission, the search area coverage is an essential part. A high explored coverage rate means the designed system is more likely to find the target, such as a victim or trapped person.

\paragraph{Cost Time}

Time is the crucial metric within the SAR context. In fire disasters a few minutes can make the difference between finding a person dead or alive.

\paragraph{Unavailable GPS Signal}

Use of GPS signals is the main approach for determining accurate positions of autonomous UAVs.
However, GPS signals can be limited or unavailable in urban canyons and tunnels. There are also obstacles which can obstruct GPS signals.

\paragraph{Unstable Communication Connection}

The system needs to be robust to an unstable communication network, due to interference or natural signal blockage.

\paragraph{Collision Free}

Challenges in cluttered space determines size UAVs smaller than $30$ cm in diameter to be useful. 

This paper focuses on a scalable, decentralised multi-UAV system for navigation, exploration and target finding in a GPS-denied and cluttered environment under uncertainty. The SAR scenarios in this paper are based on an indoor area which contains pillars and lounges. The UAVs are deployed from an empty space on the other side of entry of the hall, shown in Fig.~\ref{simen}. UAVs need to navigate, search and find a unknown target represented by an augmented reality (AR) tag attached on the ground, via using a downward-facing camera. A map of the inner space is known to the multi-UAV system. The mission is to find the target and send its position through the network. In the mission, an unstable and noisy communication network (e.g. WiFi or radio) is also considered. Furthermore, avoidance of collisions with obstacles and other UAVs is also required. It is assumed that there are no changes in flight conditions and there is no wind that will cause a disturbance of UAV flight performance. Ground level is at sea level. It is also assumed that each UAV only has access to the computational capacity of its own on-board processor.

\section{System Architecture} \label{Sysarch}

\begin{figure}[!bt]
\centering
\includegraphics[width=\columnwidth]{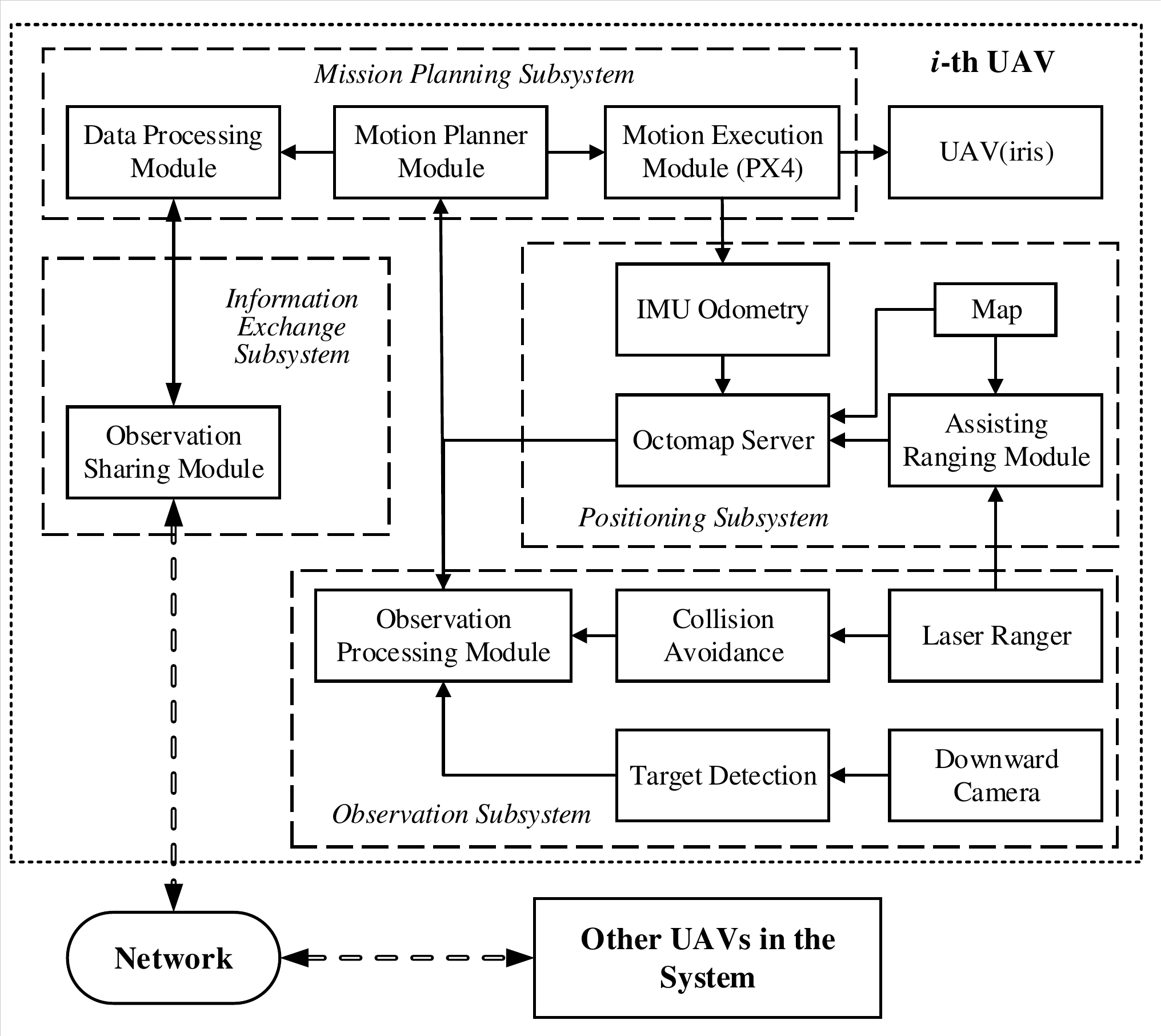}
\caption{Multi-UAV system architecture.}
\label{sysa}
\end{figure}

In order to maintain the vital advantages of a decentralised system, such as scalability and reliability, the whole multi-UAV system consists of several identical single-UAV systems which have a design taking into account the concepts of modularity and versatility, as shown in Fig.~\ref{sysa}. Each UAV system contains six modules grouped into four subsystems: Mission Planning, Positioning, Observation, and Information Exchange. The Mission Planning subsystem includes the modules for motion planner, motion execution and data processing. The Positioning subsystem can have various combinations depending on the application. In this paper, the Positioning subsystem consists of an inertial odometry module and a range sensor module. The Observation subsystem contains a target detection module and an observation processing module. The Information Exchange subsystem consists of a communication network and an observation sharing module.

\subsection{Motion Planner Module}

The motion planner module is one key part of the multi-UAV system. Its primary goal is to solve the optimal policy $\displaystyle (\pi^{*}) $ and to select an action $\displaystyle (A_{i}) $ in order to accomplish the mission. A detailed description of the decentralised POMDP used is presented in Section~\ref{Formu}. In this paper, the motion planner solver is based on the TAPIR open source software~\cite{Klimenko2014}, which is a fast online solver for large and continuous state space POMDP by using the Adaptive Belief Tree (ABT) algorithm~\cite{Kaelbling1998}.  TAPIR was formulated as a motion planner solver, using a C++ implementation. (\url{https://github.com/rdl-algorithm/tapir})

\subsection{Motion Execution Module}

After a set of actions is defined in the system formulation, the chosen action $\displaystyle (A_{i}) $ by the motion planner solver needs an autonomous UAV flight control system for manoeuvring. PX4, an open source software developed by the Dronecode Project \cite{Meier2015}, is widely used as the flight control unit (FCU) in several UAV applications \cite{Vanegas2016a,Perez-Grau2018}. The flight stack layer of the PX4 architecture contains a pipeline of feedback loop based flight controllers for various types of UAVs, such as multi-rotors, fixed-wing and VTOL, and position \& altitude estimators. A~high precision inertial measurement unit (IMU) is contained in PX4 as the inertial odometry. The PX4 also has a middleware layer, which contains the device drivers for onboard sensors, communication interfaces, and a simulation layer.

\subsection{Data Processing Module}

In this paper the map of the environment is known and divided into grids of the same size. This allows each UAV to explore a small area, which covers a grid, to find the target. With the help of this module, the UAV is able to send a observation information to other UAVs in the team in order to improve the efficiency of the whole system. There are two main functions of this module:

\begin{enumerate}[{$\bullet$}]
\item Searching and locating a new empty grid in the map by accessing the UAV current target particles in the system belief;
\item Recording the coordinates of this grid and sending it to the observation sharing module.
\end{enumerate}

\subsection{ Octomap Server with Map}

  One of the main parts of the positioning subsystem is the Octomap server which  uses the 3D occupancy voxels to represent the obstacles and boundaries within the  flying area. The map of  the environment  contains walls, doors and primary obstacles. A Gazebo simulation world is built based on the given map, and an equivalent occupancy map is also created for the Octomap server as the probability environment for the POMDP solver. The IMU odometry from the PX4 provides the basic estimate position of the UAV, the Octomap server  then sends position  estimate data to the motion planner module and  the observation processing module in the occupancy map.

\subsection{Assisting Ranging Module}

The primary purpose of this module is to calibrate the estimated position provided from the inertial odometry within the flight. There is a multi-ranger sensor that uses 5 laser sensors to measure the distance in the directions front, back, left, right, and up of the UAV. This multi-ranger sensor is a relatively precise measurement in a close range. Whit this sensor, the UAV gets a precise localisation to the motion planner when it nears a obstacle. This module is also used for a collision avoidance when the safety zone of a UAV is occupied by an obstacle. This reaction is priority than a computed action of the motion planner.

\subsection{Observation Processing Module}

The observation processing module is designed to sense the objectives, obstacles and target. The multi-ranger sensor is used for identifying obstacles in the flying area in order to avoid a collision between the UAV and the obstacles. In order to avoid UAV-to-UAV collision, this module always monitors other UAVs flight altitudes shared by the communication network  (in this paper, WiFi is used as the connection between the UAVs). The module generates a observational input to the motion planner to avoid the UAV within same altitude with its peers. The downward-facing monocular camera is used for recognising the ground target via an AR tag on the top of the target. This allows the system to calculate the position and orientation of the target in the global frame based on the UAV pose and the AR tag position in the image frame. Finally, the observation processor collects data from various sensors and combines all the shared observation in the network with its observation to update the motion planner.

\subsection{Observation Sharing Module}

This module is the critical part of the whole system. With an effective observation sharing module, the multi-UAV system performs to cooperatively accomplish the mission,
instead of each UAV working independently. The module obtains the data from the solver and simplifies the local observations to a message and frequently publishes the message through the network. A typical message contains three parts of observation: target detection status \& localisation, unified ID of the latest explored grid with this UAV ID, and all explored grids' ID. This kind of structure of the published message is convenient for monitoring and updating the observation.

\section{Problem Formulation} \label{Formu}

As described in Section~\ref{Bag}, due to the complexity of the formal Dec-POMDP and its many extensions, it is unaffordable for a onboard computer to solve the entire problem, even using state-of-the-art Dec-POMDP solving algorithms. In this paper, an observation-oriented Dec-POMDP (Ob-Dec-POMDP) model was therefore extended and modified and used to formulate and solve the problem described in Section~\ref{ProDes}. This Ob-Dec-POMDP model is expanded from our previous work \cite{Zhu2020}, which maintains the scalability and cooperation as a MAS without an intractable computational load for the current onboard processor, as a compromise and combination between the theoretical mathematic model and the real-wold implementation. The primary purpose of this Ob-Dec-POMDP formulation is to accomplish a comprehensive mission as an efficient cooperative MAS within fast reaction time in cluttered environments. The Ob-Dec-POMDP can be described as a tuple \mbox{$\displaystyle < I,S,A,T,\{\mathit{\Omega_{l}},\mathit{\Omega_{g}}\},O,R,\gamma,h > $}.

\subsection{State Variables $(S)$}

As $\displaystyle I$ represents the finite set of UAVs and can be used as a identification in the multi-UAV system, the state of each UAV is described as a tuple $\displaystyle S = < i, P_{r}, P_{t}, V_{m}, d_{c}, t_{f}> $, where $\displaystyle i $ is the identification of this UAV,  $\displaystyle P_{r} = \{x_{r} ,y_{r} ,z_{r} ,\psi_{r} \} $ are the position and orientation of this UAV based on the world coordinates,  $\displaystyle P_{t} =\{x_{t} ,y_{t} ,z_{t} ,\psi_{t} \} $ are  the position and orientation of the target in the world coordinates, $\displaystyle V_{m} = \{v_{x} ,v_{y} ,v_{z} ,v_{\psi} \}$ are the flight velocities of UAV on longitudinal, lateral, vertical, and yaw axises on the UAV body frame, then $\displaystyle d_{c} = \{ d_{xc} ,d_{yc} ,d_{zc} \} $ are the safety distance from obstacles on the UAV local frame, and the last part $\displaystyle t_{f} $ is the flight manoeuvre time for each UAV.

\subsection{Actions $\displaystyle (A)$}

The set of actions $\displaystyle(A)$ in this formulation contains several actions: Forward, Backward, Go-left, Go-right, Go-up, Go-down, Turn-left, Turn-right and Hovering, where:

\begin{enumerate}[{$\bullet$}]
\item Forward: flight forward with a distance $\displaystyle v_{x} \cdot t_{f} $ on the UAV local coordinates;
\item Backward: flight backward $\displaystyle - v_{x} \cdot t_{f} / 2 $ to avoid a front obstacle;
\item Go-left \& Go-right: flight to left or right $\displaystyle v_{y} \cdot t_{f} $ from a lateral approaching UAV or obstacles; 
\item Go-up \& Go-down: changing the altitude to avoid a obstacle or other UAVs and for a different field of view (FOV) of the downward-facing camera;
\item Turn-left \& Turn-right: change the yaw angle of the UAV by $\displaystyle v_{\psi} \cdot t_{f} $;
\item Hovering: holding the UAV current pose for more detailed observation.
\end{enumerate}

\noindent
In this paper, the actions $\displaystyle (A)$ are divided into two categories based on its purpose: the first one is the search actions (Forward, Turn-left, Turn-right, Hovering, Go-up and Go-down) which are performed the UAV to explore the scenarios to find the target; the second is the avoidance actions (Backward, Go-left, Go-right, Go-up and Go-down) that are only activated on the collision avoidance situation, when the multi-ranger sensors detects an object or another UAV in the UAV safety zone.

\subsection{Observations ($\displaystyle \{\mathit{\Omega_{l}},\mathit{\Omega_{g}}\}$)}

In order to limit the computational cost while keeping the scalability of the system, an extended observation approach is designed and tested in real-world scenarios based simulations. The observations $\displaystyle \mathit{\Omega}$ consists of two parts:

\paragraph{Local Observation $\displaystyle (\mathit{\Omega_{l})}$} this is obtained through the local on-board sensors of each UAV, which contains the status of the target detection under uncertainty, the position of the detected target in the FOV of the downward-facing camera, the ID of the explored grid, the status of avoidance mode (keeping the safety zone of UAV), the distance from the surrounding obstacles in the map.

\paragraph{Global Observation $\displaystyle (\mathit{\Omega_{g})}$} the global observation from other UAVs are provided as the a latest observation shared message only contained the useful and simplified observation information via the system communication. A shared global observation message consists of: the status of the target detection, the global position of the target in the map with the detected UAV ID, the latest explored grid ID with the ID of the searching UAV, and the IDs of all of explored grids.

\subsection{Transition Function $\displaystyle (T)$}

The transition function $\displaystyle ( T)$ models the UAV dynamic response to motion commands that are represented as the step inputs. In order to represent the $\displaystyle i $-th UAV dynamics, a~kinematic model is described as follows:

\noindent
\begin{eqnarray}
P_{r}( t+1) & = & P_{r}( t) +T_{r}( t) \vartriangle P_{r}( t) \label{eq3} \\
\vartriangle P_{r}( t) & = & V_{m}( t) \vartriangle t+P_{wt}( t) \label{eq4}
\end{eqnarray}

\noindent
where 

\noindent
\begin{equation*}
T_{r}( t) =\begin{bmatrix}
\cos( \psi _{rt} +\psi _{wrt}) & -\sin( \psi _{rt} +\psi _{wrt}) & 0 & 0\\
\sin( \psi _{rt} +\psi _{wrt}) & \cos( \psi _{rt} +\psi _{wrt}) & 0 & 0\\
0 & 0 & 1 & 0\\
0 & 0 & 0 & 1
\end{bmatrix}
\end{equation*}

 In the above equations, $\displaystyle T_{r}( t)$ is the dynamic change matrix at time step $\displaystyle t$, $\displaystyle P_{r}( t)$ is the position of $\displaystyle i$-th UAV in the world frame at time step $\displaystyle t$  and $\displaystyle P_{wt}( t) =\{x_{wt}( t) ,y_{wt}( t) ,0 ,\psi _{wt}( t)\}$ is the pose uncertainty  during the flight. The position of each UAV is determined by calculating the change in position due to its current velocity (Eq.~\ref{eq3}), which is controlled by the motion planner module and is selected according to the commanded action. A transformation from the UAV’s frame to the world  frame is also calculated in these equations.

 The PX4 controllers  use a multi-copter control architecture, which is a standard cascaded control architecture with mixed of P and PID controllers \cite{Meier2015}. The PX4 controllers are capable  of providing a comprehensive pose output for the UAV state $\displaystyle (S_{t})$ on the world frame with a solved action input.

\subsection{Reward Function $\displaystyle (R)$}  \label{sec:reward_function}

Table~\ref{reward} lists two possible options for the reward function $\displaystyle (R)$: Option A sets equal cost for all the search actions, and Option B is designed for a more aggressive exploration e.g. less hover. As well as a positive reward when each UAV explores a new grid. The reward function at each step consists of: (i) the reward of finding the target and explored a new grid in the map, (ii) the penalty of collision with a obstacle or flight out of operational space, and (iii) the cost of each manoeuvre.

\subsection{Uncertainty}

The three main sources of uncertainty of the system consist of the localisation uncertainty of the airborne UAVs, the position uncertainty of the target, and the shared observations uncertainty in the communication.

\subsubsection{Localisation of UAVs} The localisation uncertainty for each UAV in this formulation covers two parts.
In the first part, the uncertainty in localisation of system once it is initialised (on the $\displaystyle x-y$ plane) is $\displaystyle P_{wd} =\{x_{wd} ,y_{wd} ,0 ,\psi _{wd}\}$, in which the UAVs will hover for 5 seconds to initialise sensors and network connection after take-off, represented by a Gaussian probability distribution ($\displaystyle \mu = 0.0 m $, $\displaystyle \sigma = 0.5 m $).
In the second part, the UAVs' pose uncertainty during the flight is $\displaystyle P_{wt} =\{x_{wt} ,y_{wt} ,0 ,\psi _{wt}\}$, which is described by the Gaussian distribution ($\displaystyle \mu = 0.0 m $, $\displaystyle \sigma = 0.2 m $) \&  ($\displaystyle \mu = 0.0 ^\circ $, $\displaystyle \sigma = 5.0 ^\circ $). We use a downward laser sensor to calibrate the altitude and assume all the altitude data is accurate for the system through the flight. These localisation uncertainty values are updated through the transition function.

\subsubsection{Target position} The position uncertainty of the target $\displaystyle P_{t} =\{x_{t} ,y_{t}\}$ is represented by a uniform distribution in the scenario. 

\subsubsection{Shared message} The uncertainty in shared observations is caused by noisy communications. This type of noise is modelled with a confidence factor $\displaystyle \lambda = 0.9$, which means 90\% of particles representing the target positions are removed from the current belief based on the shared message of new explored grid by another UAV. Moreover, there is a double-check mechanism used for two latest shared messages to confirm the data in the messages is correct and true. These observation-related uncertainties are updated by the observation sharing module.

\begin{table}[!tb]
\caption{Reward Function Sets}
\label{reward}
\centering
\small
\begin{tabular}{|c|c|c|}
\hline 
\multirow{2}{*}{\textbf{State Status \& Actions}} & \multicolumn{2}{|c|}{\textbf{Reward or Cost Value}} \\
\cline{2-3} 
   & \textbf{\textit{Option A}} & \textbf{\textit{Option B}} \\
\hline
Detecting the target & 300 & 300 \\
Explored a new grid & 60 & 60 \\ \hline
Hitting an obstacle & -700 & -350 \\
Out of flying zone & -600 & -600 \\ \hline
Forward & -10 & -3 \\
Turn-left \& Turn-right & -10 & -5 \\ \hline
Go-up \& Go-down & -10 & -10 \\
Hover & -10 & -50 \\ \hline
Backward, Go-left \& Go-right$^{\mathrm{*}}$ & 400 & 400 \\
Go-up \& Go-down$^{\mathrm{*}}$ & 400 & 400 \\
\hline
\multicolumn{3}{l}{$^{\mathrm{*}}$only in the avoidance actions}
\end{tabular}
\end{table}

\section{Experiments}  \label{exprs}

\begin{figure}[!t]
\centering
\subfigure[Scenario one, with target locations P1~and~P2]{\includegraphics[width=\columnwidth]{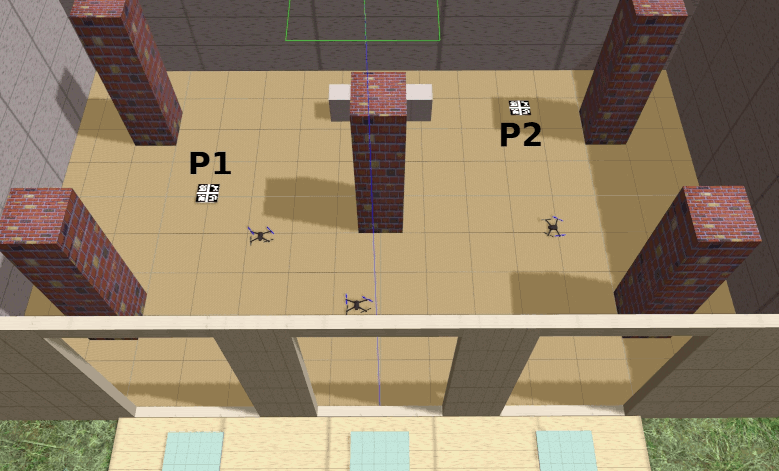}}
\\
\subfigure[Scenario two, with target locations P3~and~P4]{\includegraphics[width=\columnwidth]{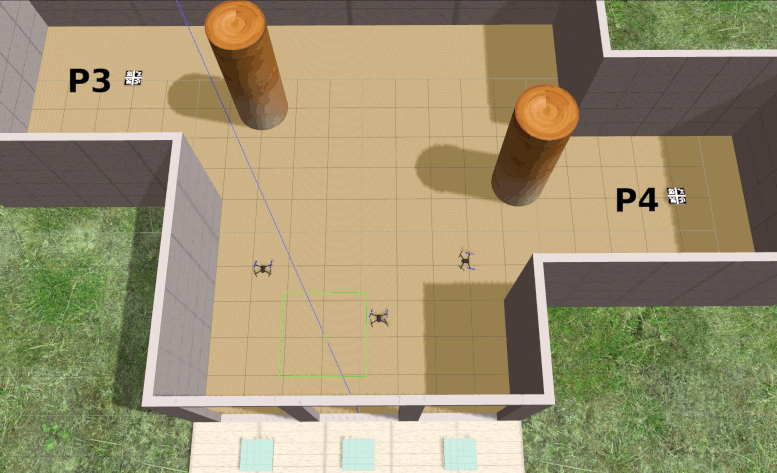}}
\caption{Simulation environments with two unique scenarios (roof not shown for illustration purposes).}
\label{simen}
\vspace{-2ex}
\end{figure}

In order to test and evaluate the performance of the proposed framework, two sets of comparative experiments have been performed.
The simulation environment consists of Robotic Operating System (ROS), Gazebo, Octomap, Rviz as visualisation tool, and PX4 at software in the loop (SITL) mode as UAV flight control unit. An off-the-shelf UAV model (3DR Iris) is used as the aerial platform in the simulations. The mission of the multi-UAV system is to explore an indoor space to find a target while avoiding obstacles in two types of complexity scenarios, as shown in Fig.~\ref{simen}. The parameters of the system were set as $ t_{f} = 1.0 ~ \mbox{sec} $, $ v_{x} = 0.8  ~ \mbox{m/s} $, $ v_{y} = 0.4 ~ \mbox{m/s} $, $ z_{t} = 0.5 ~ \mbox{m/s}$, $ v_{\psi} =  30 ^\circ/s$, $ d_{c} = \{ 0.3 \mbox{m} , 0.3 \mbox{m} , 0.3 \mbox{m} \}$. In all the experiments involving scenario one, the UAVs took off from the same locations $(-3.5,-1.0,0.2)$, $(0.0,-1.0,0.2)$ and $(3.5,-1.0,0.2)$ to find the target located in either $P1 = (-4.0,5.0,0.0)$ or $P2 = (3.5,7.5,0.0)$. While in scenario two, the UAVs took off from the same locations $(-2.5,-1.0,0.2)$, $(0.0,-1.0,0.2)$ and $(2.5,-1.0,0.2)$ to find the target located in either $P3 = (-7.0,10.0,0.0)$ or $P4 = (9.0,6.0,0.0)$. If the UAVs system can not find the target within a maximum flight time of $\displaystyle t_{\mbox{max}} = 600$ seconds, the system reports a failure to find the target.

\subsection{Experiment One}

In the first experiment, two-UAV and three-UAV simulations were implemented to search and find the target in scenario one, in order to compare the reward settings in Table~\ref{reward}, under the performance metric of {\it time to find the target} (search time). For each multi-UAV system, 40 simulations were performed, with same initial positions of UAVs and two separate target positions.

\subsection{Experiment Two}

The second experiment compares  three strategies of three-UAV simulations in the same test zones (scenario one and two):

\begin{enumerate}[{$\bullet$}]
\item Independent: this strategy sets all three UAVs as separate independent agents, where each UAV attempts to accomplish the mission by itself;
\item Divided: in this strategy, the entire search area was divided into three equal zones, and each UAV attempts to search for the target in a assigned zone without knowing any information about the other co-workers;
\item Informed: the three UAVs work jointly, and exchange messages as described in the Section~\ref{Sysarch}.
\end{enumerate}

\noindent
All the above strategies were evaluated on 40 runs, each with the same initial positions of UAVs and two separate target locations in each scenario. In this comparison, Reward Option B settings were used for all strategies to compare the search time and success rate of finding the target.

\section{Results}  \label{Res}

\begin{figure}[!tb]
\centering
\includegraphics[width=0.93\columnwidth]{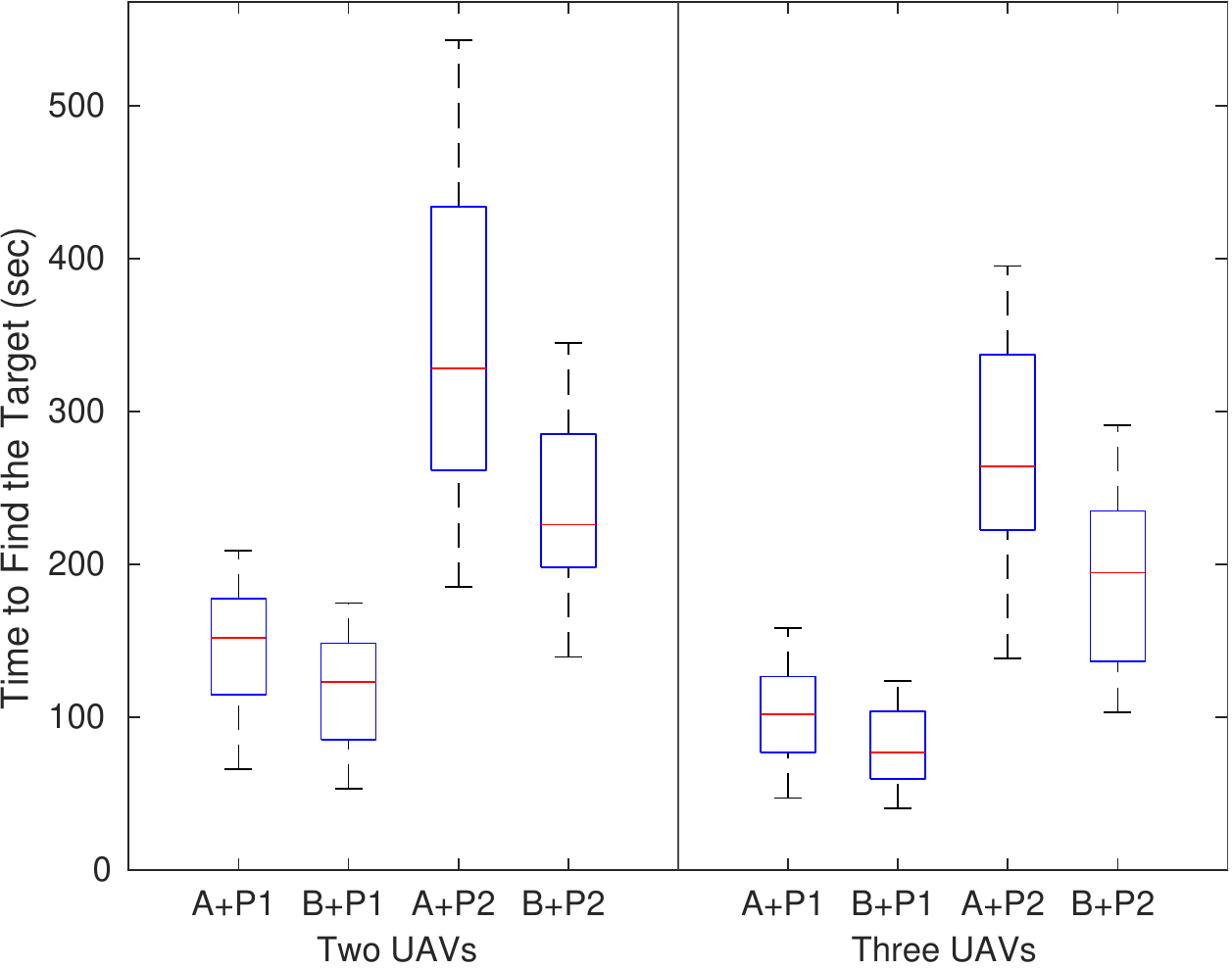}
\caption{Target finding time with various combinations of Reward options (A~and~B) and target locations (P1~and~P2).}
\label{data1}
\end{figure}

Fig.~\ref{data1} shows the results for the two reward settings (as elucidated in Section~\ref{sec:reward_function}).
Overall, it is clear that the three-UAV system spent less time to find the target than the two-UAV system with the same combination of reward options and target locations. The reduction in median target finding time was about 23\% when one UAV is added to the system. We conjecture that the performance of a multi-UAV system may be further enhanced by increasing the number of UAVs. However, the performance gains for a large group are likely to be marginal, while collision risk is likely to increase.

\begin{table}[!tb]
\renewcommand{\arraystretch}{1.3}
\caption{Target detection rate of the three strategies in the second experiment.}
\label{datatab}
\centering
\begin{tabular}{|c|c|c|c|}
\hline 
\multirow{2}{*}{\textbf{Target Location}} & \multicolumn{3}{|c|}{\textbf{Success Rate}} \\
\cline{2-4} 
   & \textbf{\textit{Independent}} & \textbf{\textit{Divided}} & \textbf{\textit{Informed}} \\
\hline 
P1 & 90\% & 97.5\% & 100\% \\
P2 & 67.5\% & 87.5\% & 95\% \\
P3 & 37.5\% & 67.5\% & 82.5\% \\
P4 & 42.5\% & 72.5\% & 87.5\% \\
\hline
\end{tabular}
\end{table}

\begin{figure}[!tb]
\hrule
\vspace{3ex}
\centering
\subfigure[Scenario one with target locations P1 and P2]{\includegraphics[width=0.93\columnwidth]{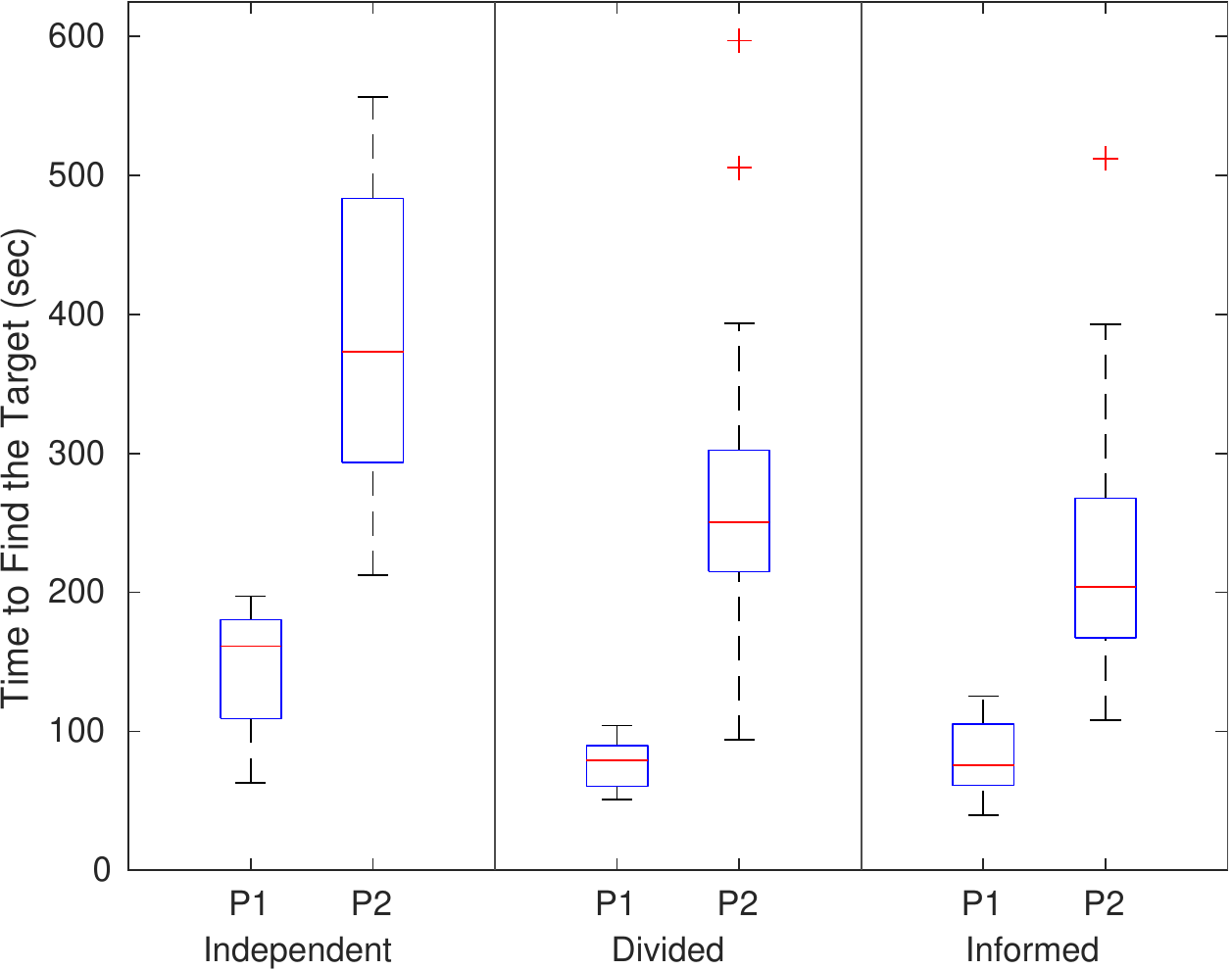}}
\\
\centering
\subfigure[Scenario two with target locations P3 and P4]{\includegraphics[width=0.93\columnwidth]{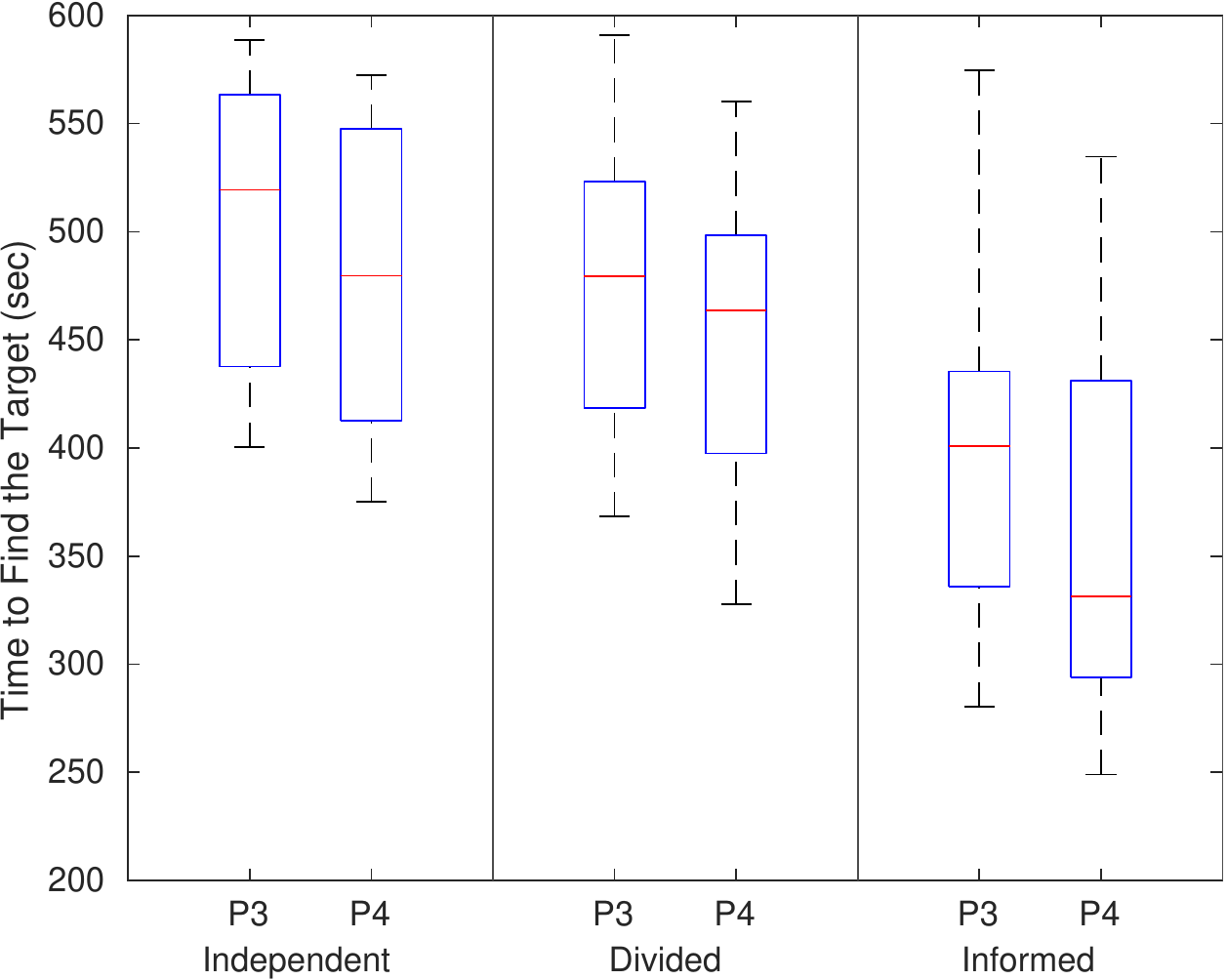}}
\caption{Target finding time for three unique strategies for four target locations (P1,~P2,~P3~and~P4).}
\label{data2}
\end{figure}

The results in Fig.~\ref{data1} also show that a tailored reward setting can enhance performance. Specifically, reward option~B provided a range of performance increase from 13.9\% to 37.4\% when compared to option~A. Due to the unequal value of cost in option~B of the reward function, the motion planner prefers the less costly action (Forward) rather than the more costly action (Hover). This type of decision helped the UAVs to explore more area and reduce the search time. The target location was a critical factor that affected performance. Comparing the target locations (P1 and P2), the median time cost increased from P1 to P2, and the interquartile range (IQR) also increased noticeably between these two locations with same reward option. This is expected, as target P2 is located in a far corner and is hence a more demanding task than target P1.

Fig.~\ref{data2} shows the effect of the three strategies on target finding time. Overall, the time taken was reduced from Independent to Informed. The IQR of the results in the Independent setting were significantly larger in contrast with the Divided and Informed settings. The Independent strategy performed poorly and its performance widely fluctuated compared to the other two categories (Divided and Informed). For the Informed strategy, the time to find target P1 was slightly longer than for Divided. We conjecture this may be due to the target being located at the border of two zones, where it can be found by two out of three UAVs. For the P2 target, the time taken in the Informed setting was lower than the Divided setting. The results for P3 and P4 indicate that the Informed strategy has a significant advantage over the Independent and Divided strategies for the more complex scenario.

Table~\ref{datatab} contrasts the target detection rates for all  strategies. For all target locations, the Independent strategy was considerably worse compared to the Divided and Informed strategies, especially for demanding places (P3 and P4) in an intricate area. However, the Informed setting kept a consistent high successful rate in all simulations, while the rate of the Divided generally dropped from 97.5\% of P1 to 72.5\% of P4. Overall, the Informed strategy obtained the highest success rates.

\section{Conclusions And Future Work} \label{Concl}

This paper presented an approach and framework for multi-UAV system search to find a target in an indoor environment without GPS/GNSS. The framework is able to consider the uncertainties present in the formulation, by using the proposed Ob-Dec-POMDP as the underlying model, in order to control computational cost without compromising the benefits stemming from a multi-agent system. The extended observation and the observation exchange subsystems are the critical and novel feature to keep the scalability and cooperation of the whole system, which can be implemented with on-board processors.

A~more specific and goal-oriented reward setting provides advantages in the decision process in comparison to an equal action cost setting. In order to navigate and localise UAVs in a GPS-denied environment, the proposed system uses a framework which combines a close range detector and a baseline map. A contrast test of various strategies is simulated to demonstrate the influence on system performance with several cooperation strategies.

The main contribution of the paper is the demonstration of the presented framework which can be implemented for a MAS and target search in a GPS-denied environment. The information sharing strategy balances the computational cost while the maintaining the system efficient and scalable.

Further avenues of research include finding and merging an affordable Simultaneous Localisation and Mapping (SLAM) system into the multi-UAV system framework. Moreover, real-world experiments which explore further constraints on the communication network systems can be performed to illustrate versatility of the designed framework.


\section*{Acknowledgements}

We thank Dr.~Hanna Kurniawati for the open source software TAPIR that was used as the online POMDP solver.



\balance

\bibliographystyle{IEEEtran}
\bibliography{IEEEabrv,ref}


\end{document}